\pdfoutput=1
\documentclass[11pt]{article}

\usepackage[final]{acl}

\usepackage{times}
\usepackage{latexsym}

\usepackage[T1]{fontenc}
\usepackage[utf8]{inputenc}

\usepackage{microtype}

\usepackage{inconsolata}

\usepackage{graphicx}
\usepackage{booktabs}
\usepackage{multirow}
\usepackage{amsmath}
\usepackage{amssymb}
\usepackage{xcolor}
\usepackage{colortbl}
\usepackage{subcaption}

\title{MIRe: Enhancing Multimodal Queries Representation via Fusion-Free Modality Interaction for Multimodal Retrieval}

\author{
 \textbf{Yeong-Joon Ju\textsuperscript{1}},
 \textbf{Ho-Joong Kim\textsuperscript{1}},
 \textbf{Seong-Whan Lee\textsuperscript{1}}
\\
 \textsuperscript{1}Department of Artificial Intelligence, Korea University
\\
 \texttt{\{yj\_ju, hojoong\_kim, sw.lee\}@korea.ac.kr}
}

\begin{document}

\maketitle

\begin{abstract}
Recent multimodal retrieval methods have endowed text-based retrievers with multimodal capabilities by utilizing pre-training strategies for visual-text alignment. They often directly fuse the two modalities for cross-reference during the alignment to understand multimodal queries. However, existing methods often overlook crucial visual information due to a text-dominant issue, which overly depends on text-driven signals. In this paper, we introduce MIRe, a retrieval framework that achieves modality interaction without fusing textual features during the alignment. Our method allows the textual query to attend to visual embeddings while not feeding text-driven signals back into the visual representations. Additionally, we construct a pre-training dataset for multimodal query retrieval by transforming concise question-answer pairs into extended passages. Our experiments demonstrate that our pre-training strategy significantly enhances the understanding of multimodal queries, resulting in strong performance across four multimodal retrieval benchmarks under zero-shot settings. Moreover, our ablation studies and analyses explicitly verify the effectiveness of our framework in mitigating the text-dominant issue. Our code is publicly available: \small\url{https://github.com/yeongjoonJu/MIRe}
\end{abstract}

\section{Introduction}

Information retrieval aims to fetch relevant information from a large collection given a user query, underpinning numerous NLP tasks such as search engines, open-domain question answering~\cite{chen2017reading,zhu2021retrieving}, and fact-checking~\cite{thorne-etal-2018-fever}.
Beyond conventional methods based on lexical similarities (e.g., TF-IDF and BM25~\cite{robertson2009probabilistic}), embedding-based retrieval methods~\cite{lee-etal-2019-latent,karpukhin-etal-2020-dense,izacard2022unsupervised,chen-etal-2024-m3} have achieved rich semantic matching by learning high-dimensional representations of queries and passages via large-scale pre-training. However, they focus on textual queries, struggling to address multimodal queries that encompass both textual and visual information.

In real-world scenarios, users often include visual references in their queries (e.g., complex objects or named entities depicted in an image), which are difficult to represent by text alone fully~\cite{liu-etal-2023-edis}. Recent multimodal retrieval methods~\cite{lin2023fine,luo-etal-2023-end,lin-etal-2024-preflmr,zhou-etal-2024-vista,zhou-etal-2024-marvel} have endowed text-based retrievers with multimodal capabilities by utilizing pre-training strategies for visual-text alignment. Most existing methods directly fuse the two modalities for cross-reference during visual-text alignment to enhance the understanding of multimodal queries. For instance, \citet{luo-etal-2023-end} and \citet{zhou-etal-2024-vista} facilitate modality interaction through early token fusion, where visual representations are prepended before passing through self-attention layers in the query encoder. Similarly, \citet{lin-etal-2024-preflmr} integrate modalities within the multimodal query using a cross-attention mechanism, where the textual query embeddings function as keys and values.

\begin{figure}
\centering
\includegraphics[width = 1.\columnwidth]{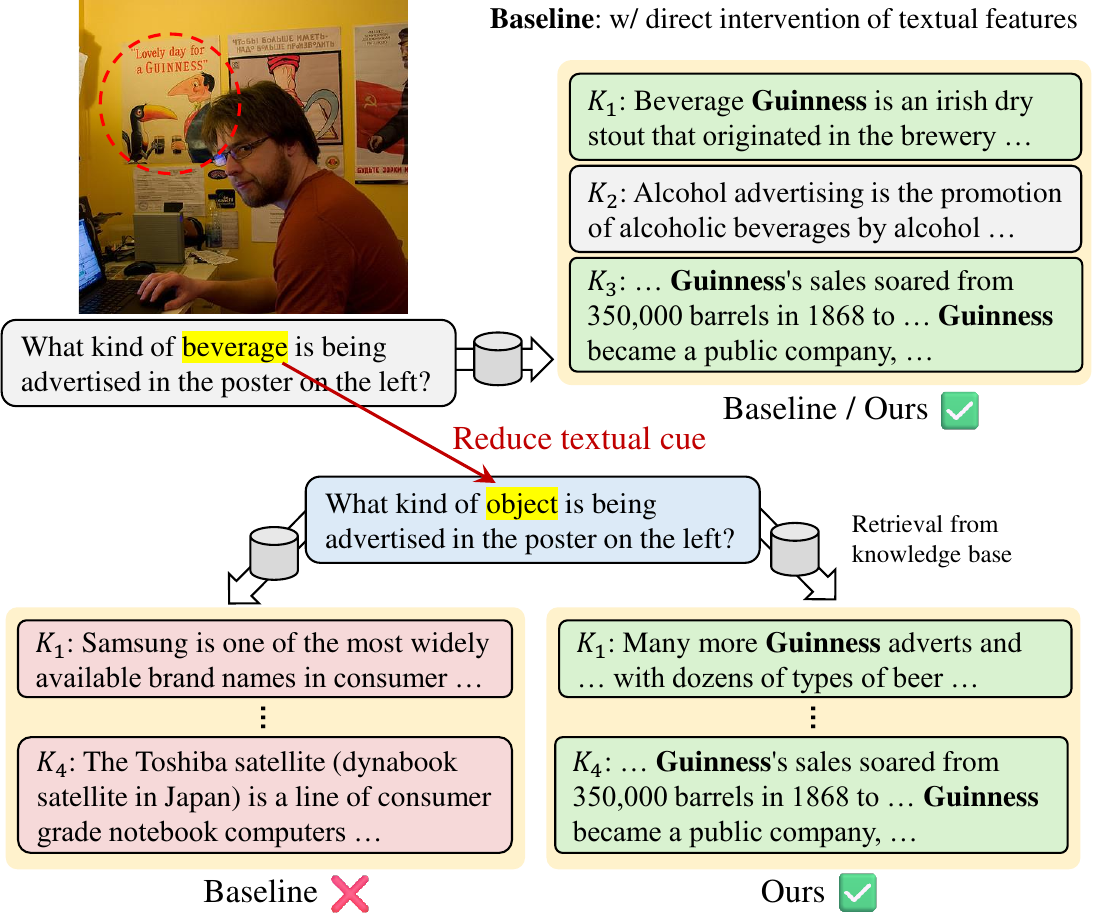}
\captionof{figure}{
    Effect of the text-dominant issue in multimodal query retrieval.}
\vspace{-8pt}
\label{fig:ex}
\end{figure}

However, these methods often overlook crucial visual information due to a text-dominant issue induced by excessive reliance on text-driven signals during the alignment stage. In this stage, the retriever over-relies on textual similarities, thereby hindering proper visual alignment. Consequently, the retriever assigns high scores to irrelevant passages when textual cues are ambiguous. Fig.~\ref{fig:ex} shows the effect of the text-dominant issue. The baseline, which is trained with a direct fusion of textual features, fails to retrieve the desired passages due to its excessive reliance on text when the textual query becomes partially ambiguous (e.g., replacing a specific term like `beverage' with a more generic word like `object'). This text-dominant issue is further amplified through pre-training datasets constructed such that pseudo-queries are extracted from passages~\cite{luo-etal-2023-end}. Datasets obtained via this approach contain text-based queries that alone are sufficient to match relevant passages. This hinders visual-text alignment by relying on the high contextual similarity between textual queries and passages, even in the absence of visual information. This issue highlights the need for a retrieval framework that leverages multimodal queries by mapping both visual and textual cues into a linguistic space, capturing complementary interactions between these modalities without over-relying on textual features alone.

To address these issues, we introduce MIRe, a retrieval framework that achieves modality interaction without fusing textual features during the alignment stage. Instead of directly merging both modalities, MIRe allows the textual query to attend to patch-level visual embeddings without feeding text-driven signals back into the visual representations. We then fuse the two modalities during the relevance scoring stage based on a late-interaction mechanism~\cite{khattab2020colbert}. This design alleviates the dependency on text-driven signals in the context of knowledge retrieval using a multimodal query. Furthermore, we construct a pre-training dataset by transforming multimodal query-response pairs into extensive passages via our response-to-passage conversion process that utilizes solely a text retrieval model. The constructed dataset requires the integration of both modalities to match a desired passage during training, enabling the model to link image understanding with complex textual queries. Our experiments demonstrate that our pre-training strategy significantly enhances multimodal query understanding for knowledge retrieval, resulting in strong performance across four multimodal retrieval benchmarks under zero-shot settings.

\section{Related Work}

Traditional methods such as TF-IDF and BM25~\cite{robertson2009probabilistic} rely on keyword matching to retrieve relevant content but often fail to capture the deeper semantics underlying queries and documents. Beyond the surface-level lexical similarities, dense retrieval methods~\cite{lee-etal-2019-latent,karpukhin-etal-2020-dense,izacard2022unsupervised,chen-etal-2024-m3,ni-etal-2022-large} leverage high-dimensional embedding models for richer semantic matching.

The transition from traditional text queries to multimodal queries has marked a significant evolution in information retrieval~\cite{luo2021weakly}. Early methods focused on converting images into textual representations, such as captions~\cite{qu2021passage, gao2022thousand} and object tags~\cite{gui-etal-2022-kat,yang2022empirical}. EnFoRe~\cite{wu-mooney-2022-entity} and DEDR~\cite{salemi2023symmetric} improve image-query representations derived from a multimodal encoder with generated entities and captions, respectively. The OVEN dataset~\cite{hu2023open} has also provided insights into multimodal entity recognition. Whereas most of these approaches utilize DPR~\cite{karpukhin-etal-2020-dense} based on a single embedding for retrieval, FLMR~\cite{lin2023fine} refines multimodal queries by incorporating RoIs and generated captions with the late-interaction mechanism. ReViz~\cite{luo-etal-2023-end} represents an end-to-end multimodal retrieval system that removes the dependency on intermediate modules by pre-training on the VL-ICT, which automatically constructs a pre-training dataset by applying the Inverse Cloze Task (ICT)~\cite{lee-etal-2019-latent} to a multimodal knowledge base. UniIR~\cite{wei2024uniir} proposes an instruction-guided multimodal retriever along with its benchmark. They design two variants of the model architecture for modality interaction: score-level fusion and feature-level fusion based on CLIP and BLIP~\cite{li2022blip}. VISTA~\cite{zhou-etal-2024-vista} introduces an in-depth fusion strategy by prepending visual tokens to the input of a text retrieval model to enhance multimodal understanding. PreFLMR~\cite{lin-etal-2024-preflmr} extends FLMR to investigate the scalability of multimodal retrievers under the late-interaction mechanism. In contrast to previous methods that rely heavily on text information within multimodal queries, we address the text-dominant issue in multimodal query representations caused by the direct intervention of textual features. We also adopt the late-interaction mechanism to fuse modalities during the scoring stage.

\section{Method}

In this section, we first define the problem of knowledge retrieval with multimodal queries. Next, we describe the architecture of our retrieval model and our data construction method.

\subsection{Problem Definition}
\label{problem}

Given a multimodal query $Q = (I, T)$, the primary objective of our retriever $\mathcal{R}$ is to retrieve a set of relevant passages $K=\{D_1, D_2, \dots, D_n\}$ from a knowledge base $U$, where $I$ and $T$ denote an image and a textual query, respectively. Each $D_i$ corresponds to a passage of text. To achieve this goal, $\mathcal{R}$ should encode the multi-modal query $Q$, integrating both the image and text modalities.

\subsection{Background: Late Interaction in Retrieval}
\label{sec:lateinteraction}
Late interaction~\cite{khattab2020colbert} is a retrieval strategy that preserves token-level embeddings for both queries and passages, enabling more fine-grained matching compared to single-vector retrieval. This mechanism defers the aggregation of embeddings to the scoring phase, retaining token-level signals. The retrieval model generates a set of low-dimensional embeddings $E = \{e_1,...,e_l\}$ for tokens in both the query and the passage. Then, the final relevance score between query embeddings $E_Q$ and document embeddings $E_D$ is computed via the following MaxSim operation:
\begin{equation} \label{eq:maxsim} 
r_{Q,D} = \sum_{i=1}^{l_Q} \max_{j=1}^{l_D} \left( E_Q \cdot E_D^T \right), 
\end{equation}
where $l_Q$ and $l_D$ denote the number of tokens in the query and the document, respectively. Each query token is matched with its most relevant document token. In our MIRe framework, we extend this mechanism to handle retrieval with multimodal queries. Our rationale for this adoption is to mitigate the overemphasis on textual features during alignment by maintaining distinct representations for each modality.

\subsection{Model Architecture}
We detail our model architecture, focusing on how it integrates visual and textual features for multimodal query retrieval.

\noindent\textbf{Textual Embeddings.} We employ a pre-trained text retriever $\mathcal{R}_T$ to encode the input textual query $T$ and passage $D$, utilizing multi-vector representations under the late-interaction mechanism. The text encoder generates token-level embeddings $E_t \in \mathbb{R}^{l_t \times d_t}$, where $l_t$ denotes the number of tokens in $T$ and $d_t$ represents the embedding dimension.

\noindent\textbf{Visual Embeddings.} We use ViT~\cite{dosovitskiy2020image} to encode image $I$. We adopt two kinds of visual embeddings: (1) global embeddings $V_g$ derived from the CLS token, representing the overall content of the image, and (2) token-level embeddings $V_m$ extracted from the penultimate layer of ViT, representing individual patches of the image. The global embedding $V_g\in \mathbb{R}^{d_v}$ is directly projected into the latent space of the text retriever $\mathcal{R}_T$ via a two-layer perception, producing embedding with dimension of $\mathbb{R}^{l_g\cdot d_t}$, where $l_g$ is the pre-defined number of tokens. Subsequently, the projected $V_g$ is reshaped into token-level embeddings $E_{g}\in \mathbb{R}^{l_g\times d_t}$.

\begin{figure}
\centering
\includegraphics[width = 1.\columnwidth]{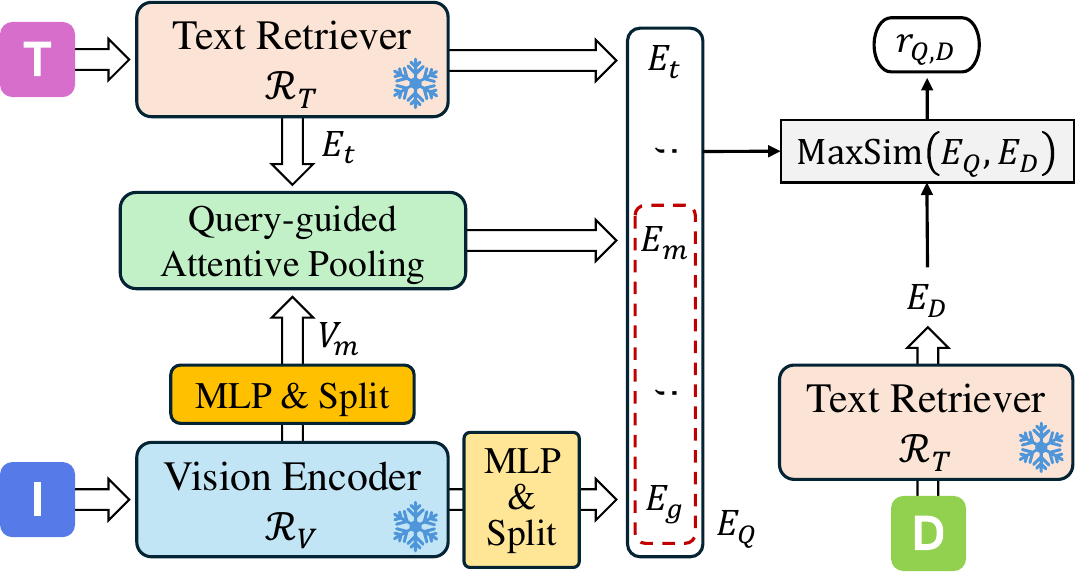}
% \vspace{-8pt}
\captionof{figure}{
    \textbf{Overview of the MIRe architecture.} This figure illustrates the interaction between the text encoder $\mathcal{R}_T$ and the vision encoder $\mathcal{R}_V$.}
\vspace{-8pt}
\label{fig:arch}
\end{figure}

\begin{figure*}[]
\centering
\includegraphics[width = 1.\linewidth]{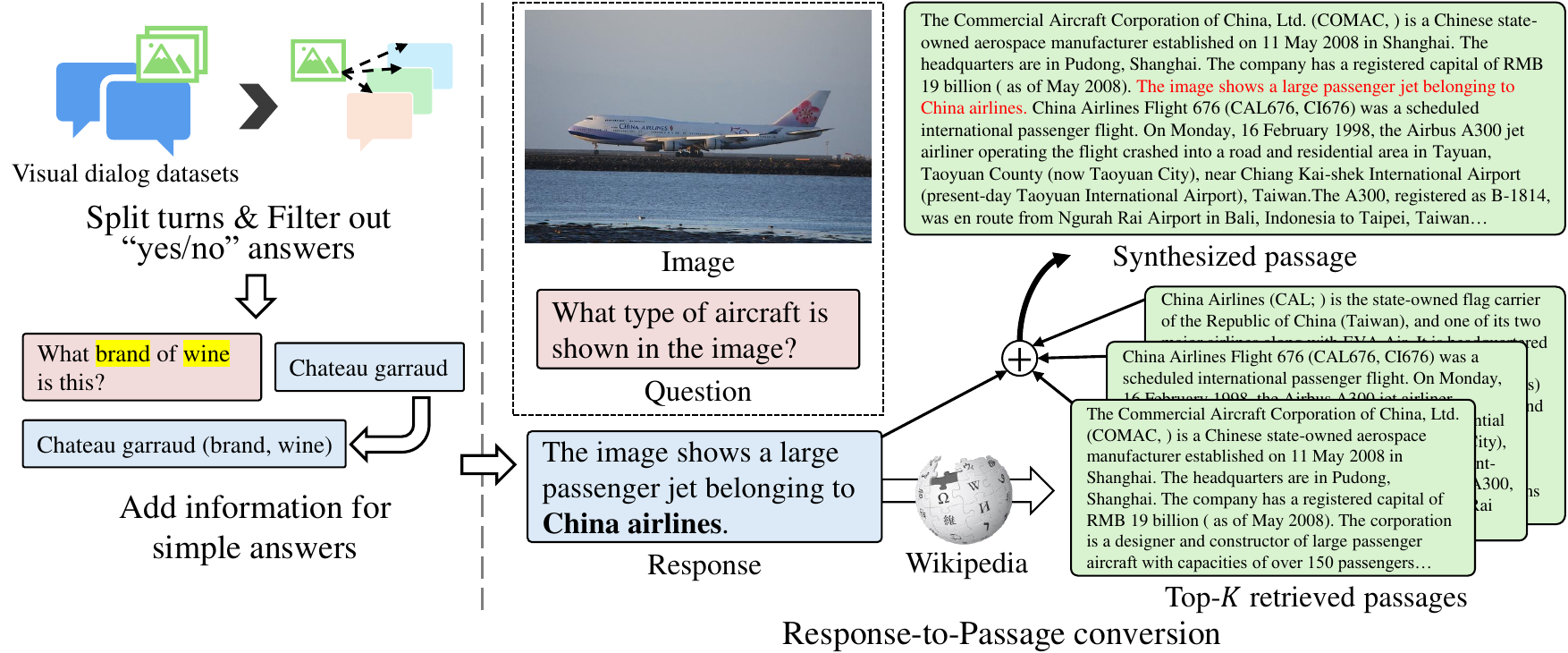}
\captionof{figure}{
    \textbf{Our data construction process.} Starting with visual dialogue datasets, our process involves two steps to convert the dialogue tasks to knowledge retrieval tasks. After preprocessing, we transform responses into a passage format by unifying the response and relevant passages retrieved from Wikipedia.}
\vspace{-8pt}
\label{fig:data}
\end{figure*}

\noindent\textbf{Query-guided Attentive Pooling.} Our architecture aims to achieve modality interaction without directly incorporating textual features in the pre-training stage for multimodal alignment, thereby mitigating the text dominance issue. To this end, we introduce a query-guided attentive pooling module employing an attention layer. This module retrieves visual information required by the textual query $T$ from $V_m\in \mathbb{R}^{l_v\times d_v}$ and then aggregates the visual information based on its relevance to tokens within the textual query, where $l_v$ denotes the number of image patches. Employing $E_t$ as query vectors, attention scores $\mathcal{A}\in\mathbb{R}^{h\times l_t \times l_v}$ are calculated as follows:
\begin{equation}
    \mathcal{A} = \text{Softmax}\left(\frac{E_t \cdot \mathcal{K}_m^\top}{\sqrt{d_t}} \right),
\end{equation}
where $\mathcal{K}_m\in\mathbb{R}^{h\times l_v \times d_t}$ denotes key vectors of $V_m$ projected by a linear layer and split into $h$ tokens for each embedding within $V_m$. Then, the attended visual output $E_m\in\mathbb{R}^{h \times d_t}$ is calculated with value vectors $\mathcal{V}_m\in\mathbb{R}^{h\times l_v\times d_t}$ of $V_m$ as follows:
\begin{equation}
    E_m = \text{Linear}\left(\frac{1}{l_t}\sum^{l_t}_i(\mathcal{A} \cdot \mathcal{V}_m)\right),
\end{equation}
where $\mathcal{V}_m$ is produced via operations identical with $\mathcal{K}_m$. Unlike the standard cross-attention mechanism, we apply mean-pooling along the sequence dimension without a residual connection, yielding $h$ visual embeddings. In this modality interaction process, we only leverage textual embeddings $E_t$ to calculate $\mathcal{A}$ as relevance scores for $T$ without direct fusion. Our module alleviates the text dominance issue by explicitly limiting textual information during visual representation alignment, as demonstrated in our empirical analyses (see Sec.~\ref{sec:discuss}).

\subsection{Dataset Construction}
We aim to train our model to comprehend images based on textual queries, thereby enabling effective multimodal query understanding. To achieve this goal, we leverage existing multimodal question-response datasets, such as visual instruction-following data and VQA data. These datasets consist of query-response pairs where each pair is associated with a single image. In each pair, the response provides a concise and image-specific answer that directly addresses the textual query. Thus, the response serves as a clear bridge between the visual content and the query, explicitly linking image understanding to the language of the query. However, despite the explicit information provided by these responses, the datasets are not directly suitable for training the retriever $\mathcal{R}$ because of the inherent difference between concise responses and more expansive passages. In practice, responses can be matched with queries without ambiguity, whereas real-world retrieval tasks demand the identification of relevant information embedded within broader documents that often contain noisy content. To bridge this gap, we transform query-response pairs into a format suitable for multimodal retrieval tasks via response-to-passage conversion, as illustrated in Fig.~\ref{fig:data}.

\noindent\textbf{Response-to-Passage Conversion.}
Let a multimodal query-response pair $S$ as follows:
\begin{equation}
    S = \{(I, T), R\},
\end{equation}
where $R$ represents the response. We first attain multiple QA pairs for a single image from samples with several turns in source datasets. We divide the response $R$ into two types: (1) detailed responses and (2) simple responses with a single word or a phrase. The simple responses often lack sufficient context to facilitate effective knowledge retrieval. Thus, we compensate simple responses with nouns extracted from the textual query $T$. Note that we filter out pairs of responses that do not contribute to knowledge-based retrieval, such as simple affirmations and negations (e.g., \emph{\textquotedblleft yes\textquotedblright} or \emph{\textquotedblleft no\textquotedblright}).

The nature of the data $S$ guarantees a high correlation between $(I,T)$ and $R$ since the responses contain information conditioned on the given multimodal query while the textual queries have restrictive information. Thus, we utilize the response $R$ to transform the response into an informative passage. From an arbitrary knowledge base $U$, we retrieve relevant passages using the response $R$ as the query. Specifically, we obtain the top-$k$ passages:
\begin{equation}
    \{D_1,D_2,\dots,D_k\} = \text{Retrieve}_{\mathcal{R}_T}(R, U, k),
\end{equation}
where $\text{Retrieve}_{\mathcal{R}_T}$ denotes the retrieval function that returns the top $k$ relevant passages from the knowledge base $U$ based on the query $R$ using the text retriever $\mathcal{R}_T$. To maintain contextual relevance with the multimodal query $(I, T)$, we then augment the response $R$ by combining it with the retrieved passages:
\begin{equation}
    R' = [D_1 ; R ; D_2 ; \dots ; D_k]\text{.}
\end{equation}

This conversion strategy yields training data that more closely mimic the complexity and noise of real-world documents. Consequently, the retriever is exposed to more challenging and realistic scenarios during training, enabling it to effectively integrate visual cues and ultimately achieve more robust retrieval performance.

\subsection{Training and Inference}
We deal with passages including the golden answers to a given question $Q$ as relevant passages $K$. To train our model, we employ in-batch negative sampling, which treats all passages in a training batch except for a passage $D$ belonging to $K$ as negative passages $\bar{K}$ for $Q$. We optimize our model by minimizing the following contrastive loss $\mathcal{L}_{CL}$ over the dataset $\mathcal{D}$:
\begin{equation}
\small
    \mathcal{L}_{CL}=-\sum_{\mathcal{D}} \log {\frac{\text{exp}(r_{Q,D}/\tau)}{exp(r_{Q,D}/\tau)+\sum_{\bar{D}\in \bar{K}}{exp(r_{Q,\bar{D}}/\tau)}}},
\end{equation}
where $\tau$ is the temperature parameter that regulates the influence of penalties on negative samples. During the alignment stage, all parameters of $\mathcal{R}_T$ and $\mathcal{R}_V$ are frozen, preserving the established text retrieval performance. To focus on visual alignment, we exclude the textual embeddings $E_t$ from the final query embedding $E_Q$, using only the visual features $[E_g; E_m]$ as $E_Q$. We also integrate a subset of a multimodal knowledge base, WiT~\cite{wit}, into our dataset to enrich the world knowledge learned during alignment. Note that this addition does not affect multimodal query understanding because the dataset consists solely of pairs of an image and a passage (i.e., it does not include a textual query). For such data, we simply assign dummy prompts for multimodal queries (e.g., What is the core object or subject shown here?). We discuss this integration in Sec.~\ref{sec:discuss} in detail.

After the alignment stage, we add textual embeddings $E_t$ to $E_Q$ when training on downstream tasks and the inference stage. For efficient retrieval, all passages within knowledge base $U$ are pre-indexed using PLAID~\cite{santhanam2022plaid}, identical to ColBERTv2~\cite{santhanam-etal-2022-colbertv2}. 

\section{Experiments}

\begin{table*}[!t]
\centering
\resizebox{1.\textwidth}{!}{%
\begin{tabular}{lcccccccccccc}
\toprule
\multirow{2}{*}{Model} & \multicolumn{3}{c}{OKVQA-GS} & \multicolumn{3}{c}{OKVQA-WK11M} & \multicolumn{3}{c}{ReMuQ} & \multicolumn{3}{c}{E-VQA} \\ \cmidrule(lr){2-4} \cmidrule(lr){5-7} \cmidrule(lr){8-10} \cmidrule(lr){11-13}
                       & MRR@5    & PR@5    & PR@10        & MRR@5    & R@5      & R@10            & MRR@5   & R@5    & R@10         & MRR@5  &  PR@5    & PR@10         \\ \midrule
CLIP~\cite{radford2021learning}                   & 19.08    & 34.54   & 50.48    & 16.45   & 29.81    & 43.0   & 0.34  & 0.78  & 1.36   &   -   &   -   &   -   \\
FLMR~\cite{lin2023fine}  & 38.15  &   57.25  &  69.42  &  32.56  & 50.61  & 62.58  &  66.67  & 72.10  & 74.95 & 29.97 & 42.0 & 50.75     \\
ReViz~\cite{luo-etal-2023-end}   & 45.77   &  64.05  &  75.39   & 44.03    & 62.43   &  73.44  & 23.61  & 39.43  & 46.77 & - & - & -  \\
UniIR~\cite{wei2024uniir} & 53.27   &  73.94  &  84.19   & -    & -   &  -  & 79.15  & 84.34  & 86.62 & 31.59 & 44.21 & 55.31  \\
VISTA~\cite{zhou-etal-2024-vista} & 55.33  &  72.83  &  81.61  & - & -  & - &  78.32  & 84.21  & 87.03   & 33.9 & 47.73 & 56.72  \\ 
PreFLMR\textsuperscript{\dag}~\cite{lin-etal-2024-preflmr} & 59.38  &  76.83  &  84.34  & 45.68  & 63.85 &  73.64  &  52.27  & 54.31  & 55.06  & 30.92 & 41.71 & 49.44  \\ \midrule
MIRe       & \textbf{63.03}   & \textbf{80.48}   & \textbf{88.15}  & \textbf{51.15} & \textbf{70.71} & \textbf{81.25}   & \textbf{83.06}     & \textbf{86.84}  & \textbf{88.56} & \textbf{41.88} & \textbf{54.24} & \textbf{61.01} \\
\rowcolor{gray!9}\text{ }\text{ }\textit{w/ ViT-large}        & 63.17   & 81.13  & 88.72  & 50.64 & 69.92 & 80.18  & 82.56  & 86.48  & 88.17 & 44.92 & 57.65 & 64.40 \\
\bottomrule
\end{tabular}%
}
%}
\caption{\textbf{Zero-shot performance of MIRe and comparison methods.} Note that FLMR was only pre-trained on the WiT dataset. PreFLMR\textsuperscript{\dag} were trained using our dataset and experimental settings. Bold indicates the highest performance.}
\vspace{-8pt}
\label{tab:zeroshot}
\end{table*}

\subsection{Setup}

\noindent\textbf{Benchmarks.} We employ four benchmarks for knowledge retrieval with multimodal queries: two variants of OK-VQA~\cite{marino2019ok}, ReMuQ~\cite{luo-etal-2023-end}, and E-VQA~\cite{mensink2023encyclopedic}. For OK-VQA, we use two versions based on different knowledge bases: OKVQA-GS, a corpus collected using Google search API as introduced in \citet{luo2021weakly}, and OKVQA-WK11M, a corpus containing 11 million Wikipedia passages compiled by \citet{qu2020open}. 

\noindent\textbf{Metrics.}
We evaluate retrieval performance using Mean Reciprocal Rank at 5 (MRR@5), Recall@$k$ (R@$k$), and Pseudo Recall@$k$ (PR@$k$) across four benchmarks. MRR@5 measures the ranking quality of the first relevant passage. For OKVQA-GS and E-VQA, which do not provide explicit ground-truth passages, we compute PR@5 by checking whether retrieved documents contain the correct answer. For OKVQA-WK11M and ReMuQ, we evaluate R@$k$ by verifying whether the target passages appear in the top-$k$ results.
%In contrast, for OKVQA-WK11M and ReMuQ, we evaluate R@$k$ by verifying whether the target passages appear in the top-$k$ results.

\noindent\textbf{Implementation Details.}
Our pre-training dataset is synthesized from three visual instruction datasets~\cite{zhang2023llavar,wang2023see,Liu_2024_CVPR} and two VQA datasets~\cite{singh2019towards,biten2019scene}, resulting in 1.35 million QA pairs, each paired with an image after preprocessing. We sampled to have no more than 12 question-response pairs per image. For the response-to-passage conversion, we utilize 6 million Wikipedia articles released by \citet{chen-etal-2023-pre-trained} as our data pool. We retrieve three candidate passages for each response using ColBERTv2, trained with the MS MARCO Passage Ranking task~\cite{nguyen2016ms}. Each passage is truncated to three sentences, and the response is inserted between the first and second passages to ensure contextual consistency. We also added 0.5 million pairs randomly sampled from WiT.

For our base model, we adopt CLIP ViT-base~\cite{radford2021learning} as a vision encoder and ColBERTv2 as a text encoder based on BERT-base~\cite{devlin-etal-2019-bert}. The number of tokens for visual embeddings $E_g$ and $E_m$ are set to 16 and 12, respectively. The value for $E_m$ is determined by the number of heads $h$ in the interaction module. The dimension of the final embeddings $d_t$ is set to 128, consistent with the text encoder. Our base model has 211M parameters. Further implementation details are provided in Appendix~\ref{sec:appendix}.
% During pre-training, we freeze all parameters of vision and text encoders. All our experiments were conducted on 4 NVIDIA RTX A6000 GPUs. 

\noindent\textbf{Comparison Methods.} We benchmark our MIRe model against a diverse set of baseline models that employ pre-training stages for visual-text alignment: CLIP~\cite{radford2021learning}, FLMR~\cite{lin2023fine}, ReViz~\cite{luo-etal-2023-end}, PreFLMR~\cite{lin-etal-2024-preflmr}, and VISTA~\cite{zhou-etal-2024-vista}. \textbf{UniIR}~\cite{wei2024uniir}, which requires an explicit instruction input, is also included; we follow their protocol and use the instruction "Retrieve a passage that answers the given query about the image" during evaluation. Both FLMR and PreFLMR utilize the same vision and text encoders as our model, where FLMR was pre-trained with a subset of the WiT dataset. For direct comparison, PreFLMR was trained using the same pre-training procedure as our model, thereby highlighting the distinct advantages of our model architecture. For zero-shot (ZS) evaluation, we exclude baselines requiring external supervision for fairness, while for few-shot (FS) evaluation, models utilizing supervision datasets are included.

\subsection{Main Results}

\noindent\textbf{Zero-shot Retrieval Performance.} 
Tab.~\ref{tab:zeroshot} shows that our method achieves superior zero-shot retrieval performance across all four benchmarks, significantly outperforming the comparison models. Despite employing a two-stage training strategy and directly optimizing the vision encoder for retrieval, VISTA still underperforms relative to our approach. Even though PreFLMR was trained under the same settings as our model, it exhibits a significant performance gap compared to our model. These results validate the effectiveness of our modality interaction approach. Our method also benefits from increased model capacity. The variant employing a larger vision encoder (ViT-large) shows similar performance to the standard model, but it further outperforms the standard model in E-VQA.

\noindent\textbf{Fine-tuning on Downstream Tasks}
We further demonstrate the adaptability of our model and the effectiveness of our pre-training task by fine-tuning models on downstream tasks. Tab.~\ref{tab:finetune} demonstrates remarkable adaptability when fine-tuned on downstream tasks. On the OKVQA-GS dataset, our model substantially outperforms all state-of-the-art models. On the ReMuQ dataset, our model still delivers strong performance, showing its competitive results. It is important to note that our method achieved higher performance on ReMuQ than VISTA in the zero-shot setting, which suggests that our pre-training and modality interaction approach endow our model with strong generalization capabilities. Notably, the variant without pre-training clearly lags behind the pre-trained model, highlighting the crucial role of our pre-training task. Furthermore, employing a larger vision encoder (ViT-large) yields additional improvements on OKVQA-GS, demonstrating the scalability of our approach. Overall, these results confirm that our model not only excels in zero-shot settings but also adapts effectively to fine-tuning on downstream tasks.

\begin{table}[!t]
\centering
\resizebox{1.\columnwidth}{!}{%
\begin{tabular}{lcccc}
\toprule
\multirow{2}{*}{Model} & \multicolumn{2}{c}{OKVQA-GS} & \multicolumn{2}{c}{ReMuQ}                             \\ \cmidrule(lr){2-3} \cmidrule(lr){4-5}
                       & PR@5             & PR@10           & R@5                       & R@10                      \\ \midrule
FLMR~\cite{lin2023fine}   & 70.63           & 81.23          & \multicolumn{1}{l}{62.76} & \multicolumn{1}{l}{74.67} \\
VRR~\cite{luo-etal-2021-weakly}    & 71.5            & 81.5           & -                         & -     \\
ReViz~\cite{luo-etal-2023-end}   & 73.35           & 83.17          & 23.61                     & 39.43   \\
GeMKR~\cite{long2024generative}  & 78.6            & 86.2           & 90.3                      & 92.7    \\ 
VISTA~\cite{zhou-etal-2024-vista} & \underline{82.06}               & \underline{90.11}              & \textbf{96.3}               & \textbf{97.3}      \\ \midrule
Ours w/o Pre-training            & 74.26               &  84.07             &  92.44                    &  94.38    \\ 
Ours                   & \textbf{83.59}           & \textbf{90.59}          & \underline{94.40}                     & \underline{96.20}                     \\
\rowcolor{gray!10}
\text{ }\text{ }\textit{w/ ViT-large} & 84.66 & 91.30 & 94.38 & 96.18 \\ \bottomrule
\end{tabular}%
}
% \vspace{-8pt}
\caption{\textbf{Fine-tuning performance on two tasks.}}
\label{tab:finetune}
\vspace{-2pt}
\end{table}

\begin{table}[!t]
\resizebox{\columnwidth}{!}{%
\begin{tabular}{llcccc|c}
\toprule
\multicolumn{2}{l}{Method} & OK-GS & OK-WK & ReMuQ & E-VQA & Avg.  \\ \midrule
\multicolumn{2}{l}{Base}  & \textbf{63.03}    & \textbf{51.15}    & 83.06 & 41.88 & \textbf{59.78} \\ \midrule
\multirow{6}{*}{PT} & \text{ }\textit{w/o WiT} & 62.54    & 50.53    & 82.63 & 40.88 & 59.15 \\
 & \text{ }\textit{w/o R2P} & 60.43    & 42.93    & 81.87 & 38.13 & 55.84 \\
 & \text{ }\textit{w/ Single $T$} & 59.72  & 49.09  & 79.27 & 29.29 & 54.34 \\ \cmidrule(lr){2-7}
 & \text{ }\textit{w/ Residual} & 61.65    & 47.95    & 80.47 & \textbf{43.06} & 58.28 \\
 & \text{ }\textit{w/o $E_m$} & 60.19    & 47.23    & 81.70 & 39.01 & 57.03 \\
 & \text{ }\textit{w/ $E_t$} & 51.38 & 42.13 & 71.69 & 32.80 & 49.50 \\ \midrule
\multirow{4}{*}{IF} & \text{ }\textit{w/o $E_m$} & 60.43 & 44.13 & 85.10 & 42.4 & 58.02 \\
 & \text{ }\textit{w/o $E_g$} & 58.4 & 44.61 & \textbf{85.91} & 40.24 & 57.29 \\
 & \text{ }\textit{w/o $E_g \& E_m$} & 52.46 & 36.0 & 71.69 & 42.48 & 50.66 \\
 & \text{ }\textit{w/o $E_t$} & 36.99 & 36.68 & 2.73 & 11.39 & 21.95 \\
\bottomrule
\end{tabular}
}
% \vspace{-8pt}
\caption{\textbf{Ablation Studies.} Retrieval performance (MRR@5) in zero-shot settings across four datasets. "PT" and "IF" indicate ablations performed at the pre-training and inference stages, respectively.}
\label{tab:ablation}
\vspace{-4pt}
\end{table}

\subsection{Ablation Studies}
Our ablation studies, summarized in Tab.~\ref{tab:ablation}, reveal that each component in our framework plays a significant role in achieving robust zero-shot retrieval performance. We examine the contributions of our design from three perspectives: the dataset, model architecture during alignment, and the embeddings used at inference.

\noindent\textbf{Dataset.} In the pre-training stage (PT), omitting external knowledge from the WiT dataset causes only a slight performance drop, underscoring its supportive role (see Sec.~\ref{sec:discuss}). In contrast, training the model on original responses without applying the response-to-passage conversion (R2P) results in a substantially larger decline. These observations indicate that the R2P mechanism is essential for enhancing visual-text alignment and overall knowledge retrieval. We also investigate the effect of multiple QA pairs per image. As shown in Tab.~\ref{tab:ablation}, although sampling a single QA pair per image keeps the total number of images, this variant (w/ Single $T$) significantly degrades retrieval performance, suggesting the presence of hard-negative effects beyond simple visual-image alignment.

\noindent\textbf{Model.} We further examine how directly fusing text features during the alignment process affects performance. When we add a residual connection to our model architecture before sequential-wise pooling (\textit{w/ Residual}), we observe a performance drop, indicating a slight exacerbation of the text-dominant issue. Moreover, when text features are allowed an even more direct influence, by setting $E_Q=[E_g, E_m, E_t]$ during alignment, the performance degrades considerably.

\noindent\textbf{Embeddings $E_Q$.} At the inference stage (IF), our analysis shows that each embedding type plays a unique and complementary role. Removing either the modality-specific embedding (\textit{w/o $E_m$}) or the general embedding (\textit{w/o $E_g$}) leads to a moderate decline in performance, suggesting that both capture distinct yet essential aspects of the data. However, removing these components simultaneously causes a sharper performance drop. Notably, omitting the text embedding (\textit{w/o $E_t$}) results in severe degradation of retrieval accuracy, indicating that $E_t$ is indispensable for maintaining semantic coherence. This clear hierarchy in the impact of each embedding underscores their distinct functions and the need for their balanced integration.

\begin{figure}[!t]
\centering
\includegraphics[width = 1.\columnwidth]{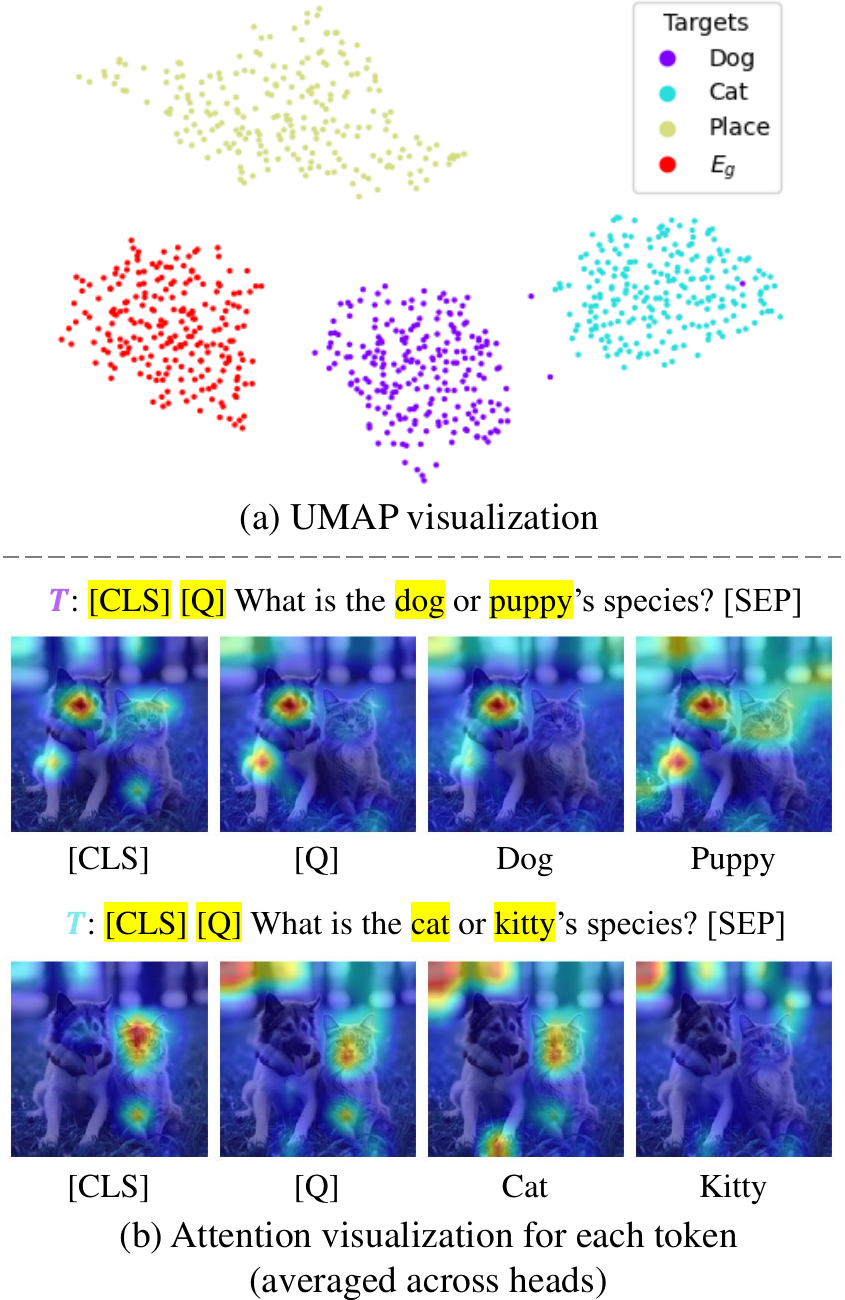}
\captionof{figure}{
    \textbf{Visualization of multimodal query processing}, illustrating the alignment between textual and visual modalities.}
\vspace{-14pt}
\label{fig:viz}
\end{figure}

\section{Discussion}
\label{sec:discuss}

\noindent\textbf{Effect of Query-guided Attentive Pooling} To demonstrate the effectiveness of MIRe in capturing modality interactions, we visualize the embeddings and attention maps of multimodal queries on a controlled dataset. We synthesized 224 images with the prompt `\textit{A dog and a cat in an image}' using Diffusion-XL~\cite{podell2024sdxl}, and conditioned the embeddings $E_m$ on three distinct textual prompts: (1) \textit{What is the dog or puppy's species?}, (2) \textit{What is the cat or kitty's species?}, and (3) \textit{Where is the place in the image?} In Fig.~\ref{fig:viz}(a), the UMAP clustering~\cite{mcinnes2018umap-software} of $E_g$ and $E_m$ illustrates MIRe effectively separates visual embeddings based on the query's intent. Additionally, Fig.~\ref{fig:viz}(b) visualizes attention patterns of our pooling module, revealing how the model attends to specific visual patches relevant to each query. These results demonstrate that MIRe enhances interactions between textual and visual modalities.

\begin{table}[!t]
\centering
\resizebox{1.\columnwidth}{!}{%
\begin{tabular}{llcccc}
\toprule
\multicolumn{2}{l}{Dataset}       & OKVQA-GS & ReMuQ & E-VQA & Infoseek \\ \midrule
\multicolumn{2}{l}{FLMR~\cite{lin2023fine}}          & 57.25    & 72.10 & 42.0  & \underline{42.93} \\ \midrule
& \textit{w/o WiT}       & \textbf{81.11} & \textbf{87.45} & 51.95 & 37.15    \\
Ours & \textit{w/ WiT (0.5M)} & \underline{80.48}  & \underline{86.84} & \textbf{54.24}  & 42.61 \\
& \textit{w/ WiT (1.0M)} & 79.63  &  86.59  & \underline{54.05} & \textbf{44.01}   \\  \bottomrule
\end{tabular}%
}
\caption{Zero-shot retrieval performance (R@5) under knowledge integration settings using WiT data.}
\vspace{-8pt}
\label{tab:info}
\end{table}

\begin{figure}[!t]
\centering
\includegraphics[width = 1.\columnwidth]{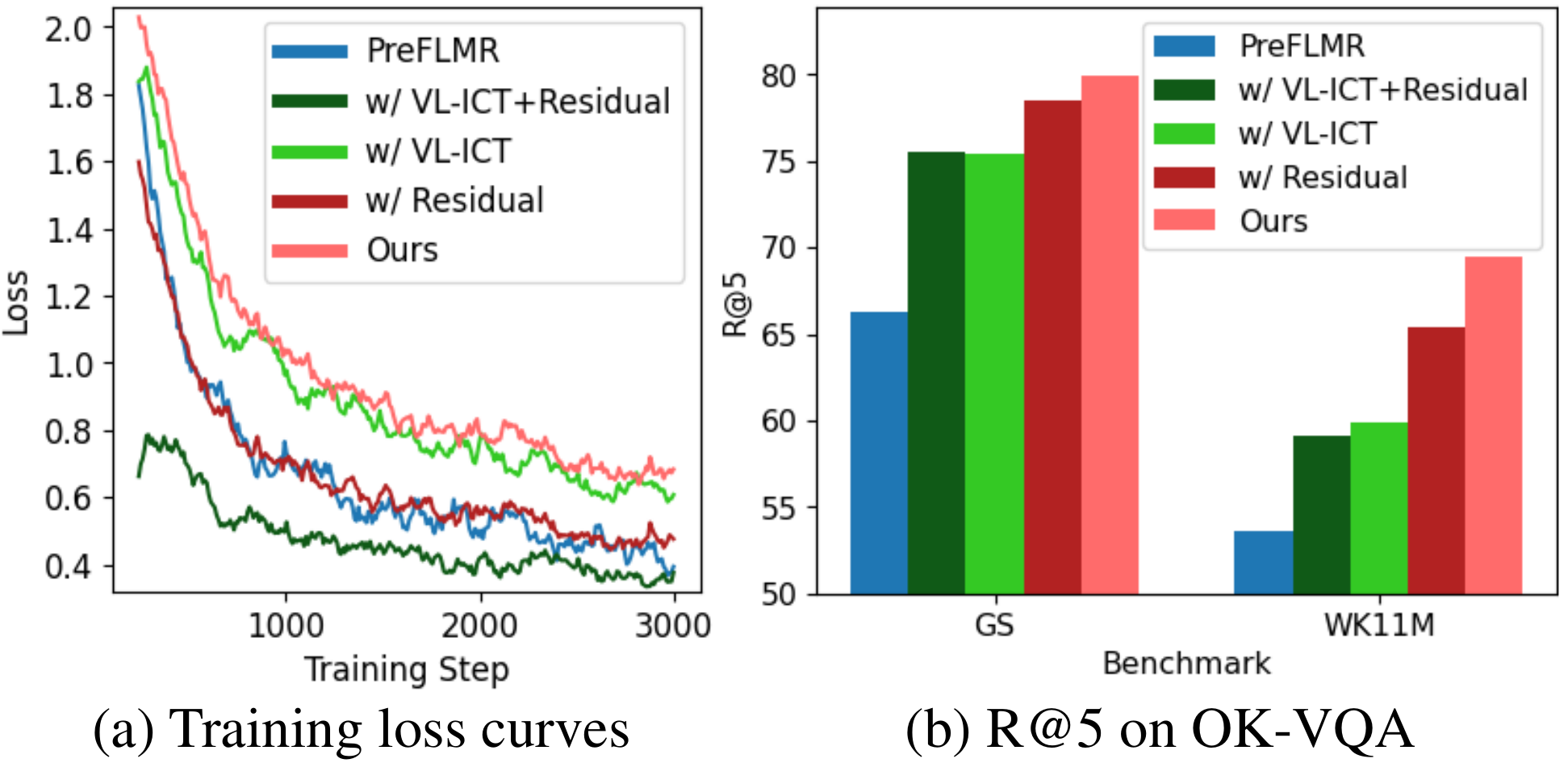}
% \vspace{-15pt}
\captionof{figure}{
    \textbf{Training convergence and retrieval performance.} All models were trained for only one epoch under the same settings.}
\vspace{-8pt}
\label{fig:ti}
\end{figure}

\begin{figure}[!t]
\centering
\includegraphics[width = 1.\columnwidth]{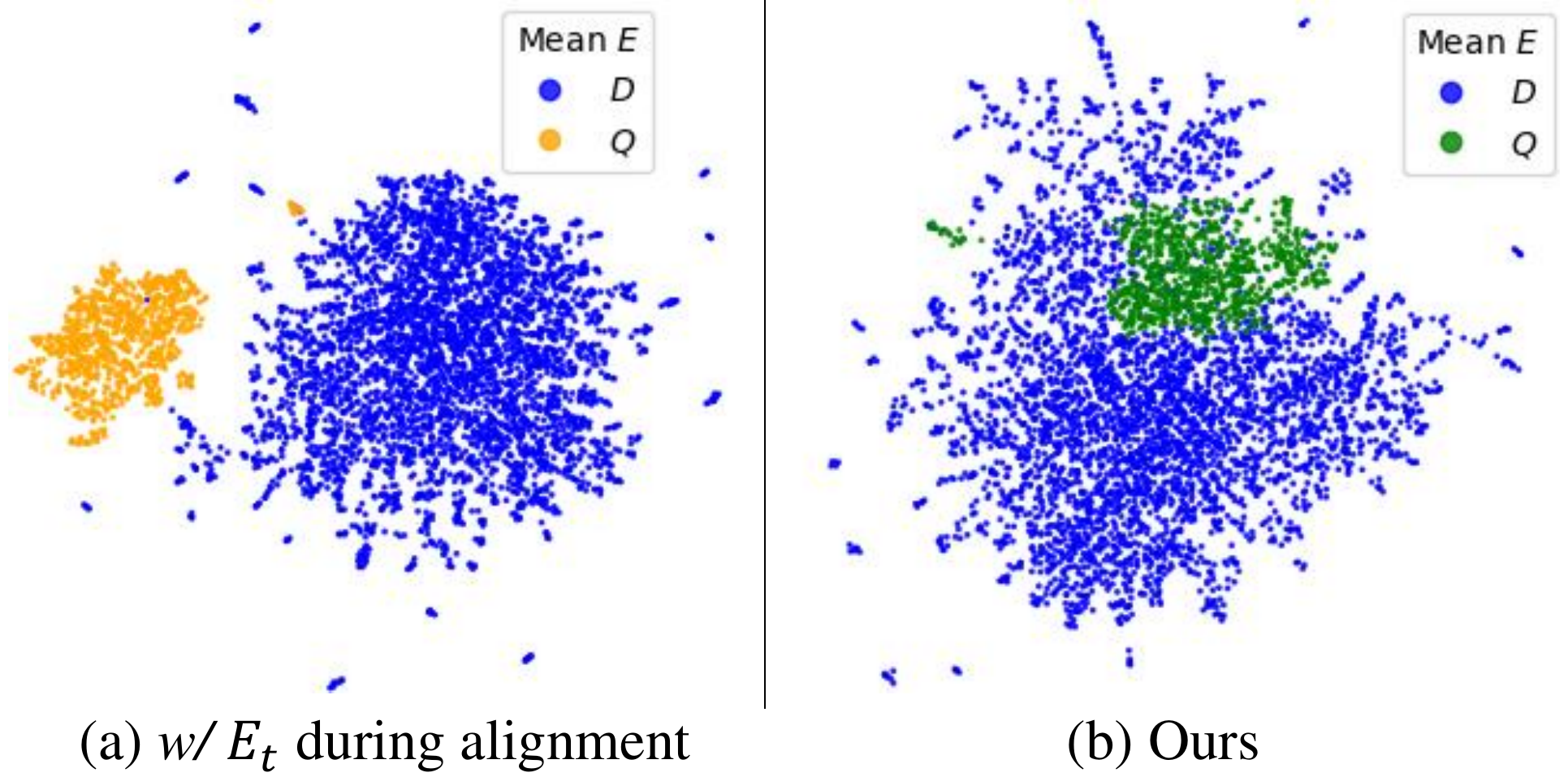}
% \vspace{-15pt}
\captionof{figure}{
    \textbf{Comparison of Embedding Distribution.} (a) with $E_t$ during alignment, where query embeddings ($Q$, orange) remain distinct from passage embeddings ($D$, blue); (b) our method, where $Q$ (green) is better integrated into the textual space.}
\vspace{-8pt}
\label{fig:align}
\end{figure}

\noindent\textbf{Effect of Knowledge Integration.}
We further assess MIRe's capacity for external knowledge integration by incorporating the WiT dataset and analyzing its effect on retrieval performance, particularly on the Infoseek dataset~\cite{chen-etal-2023-pre-trained}. As shown in Tab.~\ref{tab:info}, \textit{Ours w/o WiT} falls short on Infoseek relative to FLMR while competitively performing on other benchmarks. Notably, FLMR was learned with a subset of WiT without modality interaction. When we integrate external knowledge using 0.5 million WiT data, our model's performance on Infoseek is substantially improved to 42.61, bringing it on par with FLMR. Moreover, further increasing the WiT data to 1.0 million boosts the R@5 on Infoseek to 44.01. These findings, however, reveal that the performance gains observed on Infoseek are largely driven by its heavy reliance on external knowledge, which raises concerns about the generality of evaluation protocols that depend on such background information.

\noindent\textbf{Text-dominant Issue.} We analyze how certain pre-training strategies and model architectures exacerbate reliance on textual features during multimodal alignment. In Fig.~\ref{fig:ti}(a), both PreFLMR and \textit{w/ Residual} exhibit faster loss convergence compared to our model, suggesting that directly leveraging text features accelerates optimization. However, as shown in Fig.~\ref{fig:ti}(b), the accelerated convergence does not translate to improved performance, with PreFLMR and \textit{w/ Residual} underperforming relative to our model. The text-dominant issue is further exacerbated when using VL-ICT, introduced in ReViz, which constructs pseudo-queries from passages. Such behavior reveals the text-dominant issue, where excessive dependence on text features during alignment hinders the model's ability to fully leverage multimodal information. Fig.~\ref{fig:align} illustrates this effect by visualizing the alignment of multimodal query embeddings $Q$ with passage embeddings $D$. In (a), when $E_t$ is explicitly used during alignment $Q$ embeddings (orange) remain largely separated from the passage space, indicating poor alignment. In contrast, (b) demonstrates that our method effectively incorporates $Q$ embeddings (green) into the linguistic space, improving alignment. These results suggest that excessive reliance on text features inhibits the multimodal query embeddings from adapting properly to the passage space.

\section{Conclusion}
We introduced MIRe, a novel retrieval framework designed for multimodal query retrieval without fusing textual features during the alignment stage.
Our query-guided attentive pooling module allows textual embeddings to attend to visual patches while preventing text-driven signals from dominating the visual representations. We also constructed a pre-training dataset by converting concise question-answer pairs into extended passages, thereby exposing the model to more realistic retrieval tasks. Our extensive experiments demonstrate that MIRe consistently outperforms existing methods under both zero-shot and fine-tuned settings. Ablation studies further validate that each component of MIRe is crucial for achieving robust multimodal query retrieval.

\section{Limitations}
Despite the promising results, our work has several limitations that point to potential directions for future research. First, while our approach demonstrates strong performance across general-domain benchmarks, it remains untested in specialized domains (e.g., medical or legal documents), where multimodal content may exhibit more complex and domain-specific features. Second, we have not explored synergy with retrieval-augmented generative (RAG) frameworks, which typically prepend retrieved passages to a language model for downstream generation tasks. Although we believe our retrieval improvements would benefit RAG-based methods, in line with findings from \citet{kim2024sure} showing that stronger retrievers enhance downstream generation, fully validating our approach in a RAG pipeline is left for future work. Finally, our current data construction method focuses on retrieval from large yet homogeneous corpora; adapting the framework to more diverse or dynamically changing knowledge sources may require additional techniques to handle domain shifts or continuously updated information.

\section{Acknowledgment}
This work was supported by the Agency For Defense Development by the Korean Government(UI247035TF)

\bibliography{custom}

% \bigskip

\appendix

\section{Appendix}
\label{sec:appendix}

\begin{table*}[]
\centering
\resizebox{.85\textwidth}{!}{%
\begin{tabular}{lcccccc}
\toprule
\multirow{2}{*}{Dataset} & \multicolumn{6}{c}{Hyperparameter} \\
 & LR & \# Epochs & \# Batch per GPU & \# Global Batch & \# Warm-up & $\tau$ \\ \midrule
Pre-training (R2P) & 1e-4 & 5 & 128 & 512 & 300 & 0.3 \\
OKVQA-GS & 5e-5 & 15 & 128 & 512 & 10 & 0.8 \\
ReMuQ & 5e-5 & 5 & 128 & 512 & 10 & 0.8 \\ \bottomrule
\end{tabular}%
}
\caption{\textbf{Summary of hyperparameters utilized for training.} The LR denotes the learning rate.}
\label{tab:hparam}
\end{table*}

% For further information, we provide our code utilized in this project at the following GitHub repository:
% \url{https://anonymous.4open.science/r/MIRe3B8C}

\subsection{Training and Inference Details}
In all experiments, we train models using the AdamW optimizer~\cite{loshchilov2018decoupled} with warm-up steps on a machine with 4 RTX A6000 GPUs. We chose model checkpoints based on the validation loss. We set hyperparameters for each dataset as shown in Tab.~\ref{tab:hparam}.

\noindent \textbf{Pre-training.} We used $E_Q=[E_g;E_m]$ without $E_t$ to align visual embeddings with the linguistic space during the pre-training stage. In this stage, we only tuned the mapping network, such as a MLP layer for $E_g$ and the query-guided attentive pooling module. PreFLMR and MIRe were set with the same hyperparameters.

\noindent \textbf{Fine-tuning.} For fine-tuning our model on downstream tasks, we tuned all parameters of our model except for the vision model in all experiments. Since the parameters of the vision model are not updated during training, we cached the outputs of the vision model before training. In our setting, training one epoch for our dataset took about 20 minutes on 4 RTX A6000 GPUs, where one epoch encompasses 3625 steps. We detail statistics of benchmark datasets in Tab.~\ref{tab:stat}.

\begin{table}[t]
\centering
\resizebox{\columnwidth}{!}{%
\begin{tabular}{lccc}
\toprule
\multirow{2}{*}{Dataset} & \multicolumn{3}{c}{Size} \\
 & \#Train & \#Test & KB $U$ \\ \midrule
OKVQA-GS &  8,958 & 5,046 & 166,390 \\
OKVQA-WK11M &  - & 2,523 & 11,000,000 \\
ReMuQ & 8,418 & 3,609 & 195,387 \\
E-VQA & - & 3,750 & 51,462 \\
\bottomrule
\end{tabular}
}
\caption{\textbf{Summary of dataset statistics for evaluation.} This table presents the distribution of training and testing instances alongside the size of the knowledge bases for each dataset employed in our study. GS and KB denote the corpus collected from the Google Search API and used knowledge base, respectively.}
\label{tab:stat}
\end{table}

\begin{figure}[!t]
\centering
\includegraphics[width = .85\columnwidth]{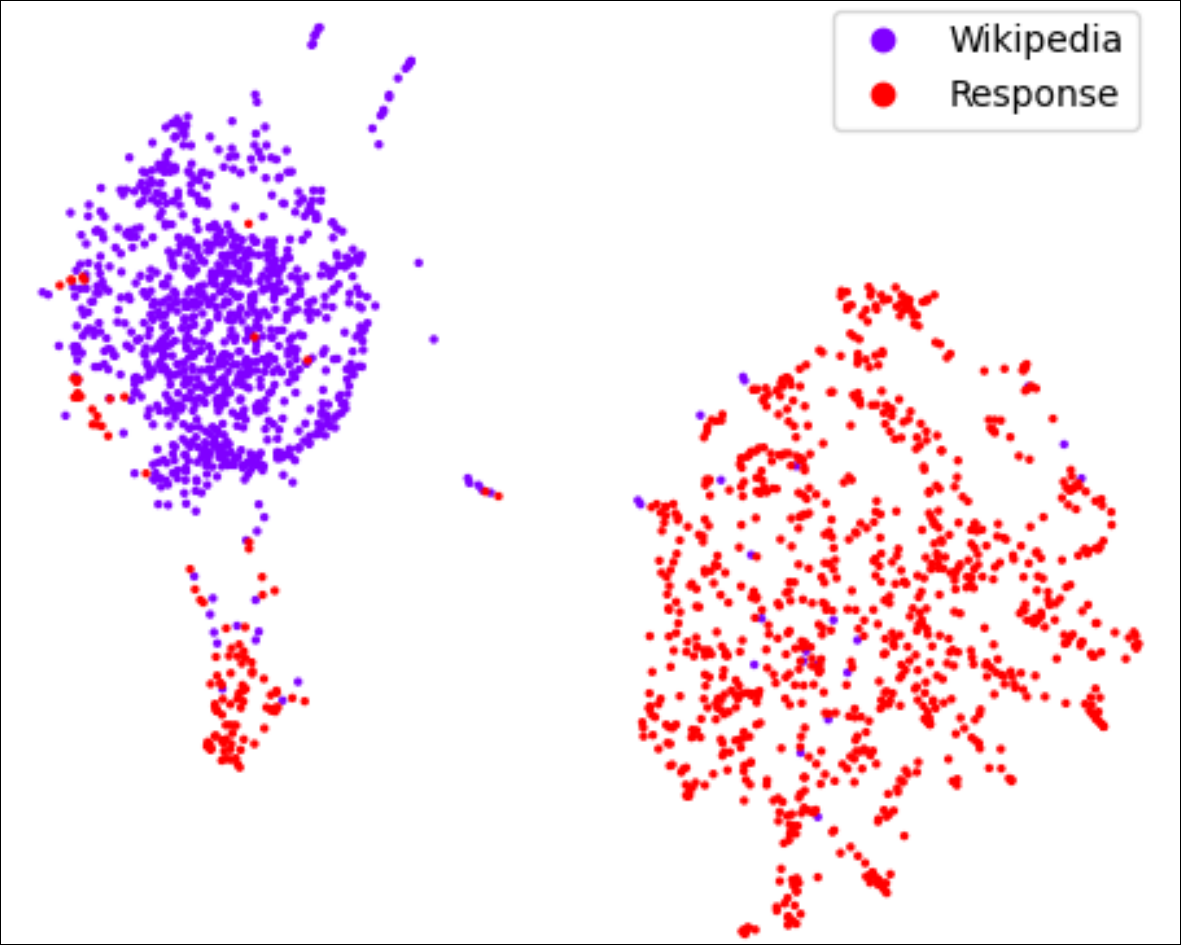}
% \vspace{-8pt}
\captionof{figure}{
    \textbf{UMAP visualization of embeddings} extracted using the Contriever model~\cite{izacard2022unsupervised}, comparing Wikipedia documents (purple) and LLaVA responses (red). The separation between clusters highlights the structural and semantic differences.}
\label{fig:dist}
% \vspace{-15pt}
\end{figure}

\begin{table}
\centering
\resizebox{\columnwidth}{!}{%
\begin{tabular}{lr}
\toprule
\textbf{Statistic} & \textbf{Counts} \\ \midrule
\# Total data & 1,356,536 \\ \midrule
\# Images & 264,262 \\ 
\# Max. queries per image & 12 \\
\# Avg. queries per image & 8.32 \\
\# Queries requiring description  & 230,877 (17.02\%) \\
\# Other types of queries & 1,125,659 (82.98\%) \\ \bottomrule
\end{tabular}
}
\caption{\textbf{Statistics of our constructed dataset.}}
\label{tab:data_statistics}
\end{table}

\begin{table*}[!t]
\centering
\resizebox{.95\textwidth}{!}{
\begin{tabular}{lcccc}
\toprule
Source Dataset & \# Data & \# Images & \# Avg. $S$ per $I$ & \# Max. $S$ per $I$ \\ \midrule
ST-VQA~\cite{biten2019scene}         & 25,154  &  18,518   &  1.36   &  7 \\
TextVQA~\cite{singh2019towards}      & 26,406  &  18,913   &  1.40   &  2 \\
LLaVAR~\cite{zhang2023llavar}        & 42,690  &  19,787   &  2.16   &  7 \\
Instruct4V~\cite{wang2023see}        & 222,711 &  26,663   &  8.35   & 12 \\
LLaVA-1.5~\cite{Liu_2024_CVPR}       & 1,017,622 & 158,429 &  6.42   & 12 \\
Subset of WiT~\cite{wit}             & 500,000 & 500,000   &  1      &  1 \\ \bottomrule
\end{tabular}
}
\caption{\textbf{Statistics of each source dataset within our dataset.} $S$ per $I$ denotes the number of queries per image.}
\label{tab:each}
\end{table*}

\begin{table*}[!t]
\centering
\resizebox{.987\textwidth}{!}{%
\begin{tabular}{lcccccccc}
\toprule
\multirow{2}{*}{Source Dataset} & \multicolumn{2}{c}{OKVQA-GS} & \multicolumn{2}{c}{OKVQA-WK11M} & \multicolumn{2}{c}{ReMuQ} & \multicolumn{2}{c}{E-VQA} \\ \cmidrule(lr){2-3}\cmidrule(lr){4-5}\cmidrule(lr){6-7}\cmidrule(lr){8-9}
 & PR@5 & PR@10 & R@5 & R@10 & R@5 & R@10 & PR@5 & PR@10 \\ \midrule
ST-VQA~\cite{biten2019scene} & 72.29 & 81.45 & 57.35 & 67.97 & 85.79 & 87.81 & 51.01 & 58.37 \\
TextVQA~\cite{singh2019towards} & 72.18 & 81.75 & 57.83 & 69.20 & 86.03 & 87.97 & 51.41 & 58.80 \\
LLaVAR~\cite{zhang2023llavar} & 73.11 & 82.62 & 60.88 & 71.90 & 86.34 & 88.39 & 51.89 & 58.75 \\
Instruct4V~\cite{wang2023see} & 78.72 & 86.88 & 65.83 & 75.90 & 86.01 & 87.81 & 52.0 & 59.25 \\
LLaVA-1.5~\cite{Liu_2024_CVPR} & 79.41 & 87.77 & 68.05 & 78.32 & 86.56 & 88.31 & \textbf{52.56} & \textbf{59.92} \\ \midrule
Total & \textbf{81.11} & \textbf{88.84} & \textbf{70.55} & \textbf{82.20} & \textbf{87.45} & \textbf{88.45} & 51.95 & 59.12 \\ \bottomrule
\end{tabular}%
}
\caption{\textbf{Zero-shot performance by each source dataset.} We apply our response-to-passage conversion process to each source dataset. Note that we did not add WiT data in this experiment.}
\label{tab:data_p}
\end{table*}

\noindent \textbf{Inference.} Passages within the knowledge base were pre-indexed, following the method established by the previous work~\cite{santhanam-etal-2022-colbertv2}. The indexing process consists of three critical steps: centroid selection, passage encoding, and index inversion. To enhance storage efficiency, embeddings were compressed to 2 bits per dimension. In the OK-VQA dataset using a corpus collected from Google search API, the retrieval time of MIRe and ColBERTv2 is approximately 0.085 seconds and 0.081 seconds per query on one RTX A6000 GPU, respectively. Thus, MIRe spends slightly more time retrieving relevant passages with multimodal queries, compared to the base text retriever.

\subsection{Details for Our Dataset}
To construct our dataset, we employ three visual instruction datasets~\cite{zhang2023llavar,wang2023see,Liu_2024_CVPR} and two VQA datasets~\cite{singh2019towards,biten2019scene}. Initially, samples were split into individual turns. We removed turns with responses shorter than 30 characters only for detailed responses. Subsequently, we edited responses containing simple affirmations (``yes", ``no") and excluded samples for tasks irrelevant to retrieval tasks (e.g., location and count), where we automatically filtered out based on specific phrases. 

\begin{figure*}[]
\centering
\includegraphics[width = 1.\textwidth]{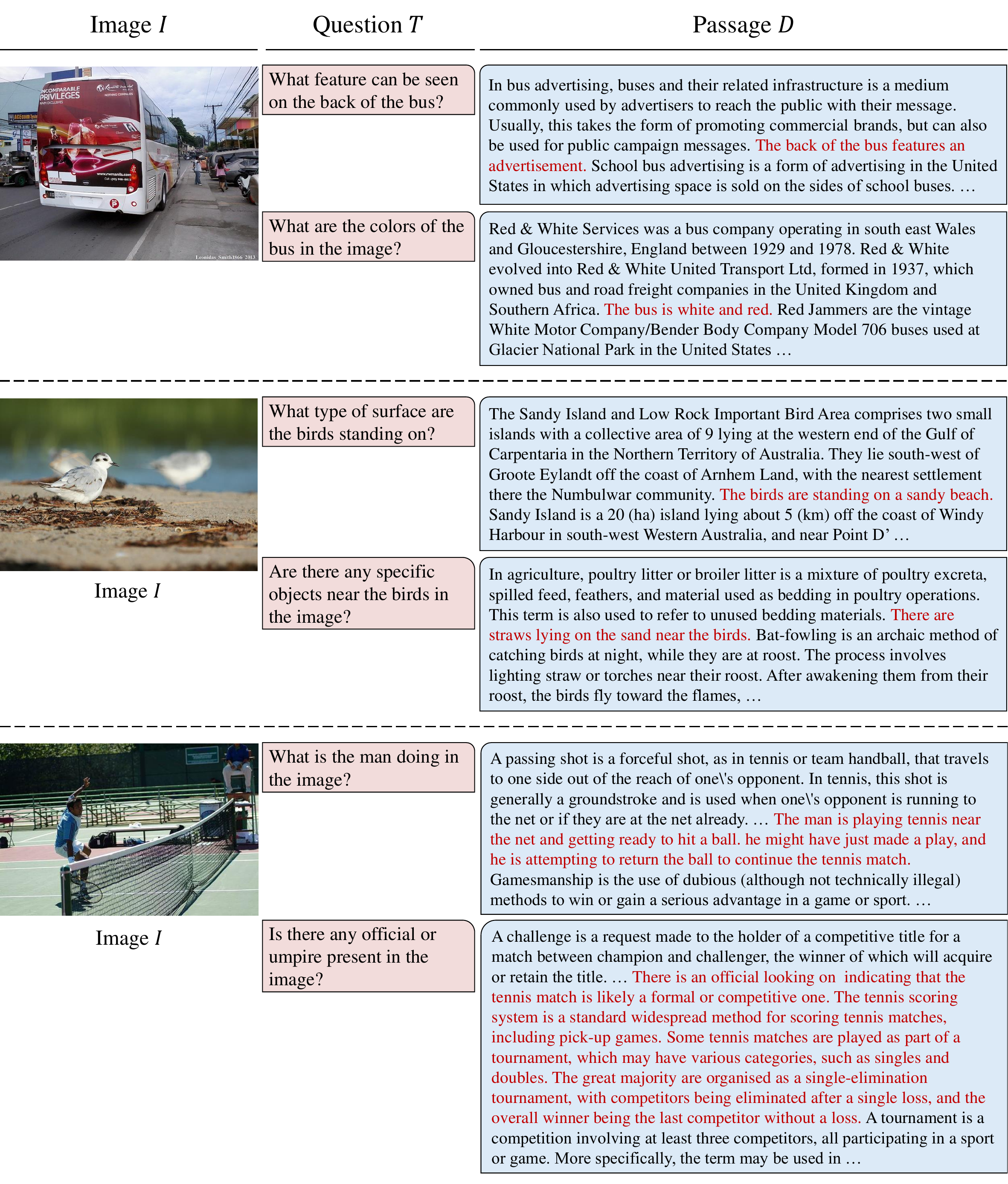}
\captionof{figure}{
    \textbf{Examples for our dataset.} The figure illustrates samples in the dataset, where the red-colored text denotes inserted responses.}
\label{fig:sample}
\end{figure*}

Fig.~\ref{fig:dist} illustrates there exists a clear distinction between the concise responses and more expansive passages, supporting our perspective. After the pre-processing, we refined the data through a response-to-passage conversion using ColBERTv2, a text retriever trained on the MS MARCO Passage Ranking task~\cite{nguyen2016ms}. Responses were converted into passages using a pool of 6 million Wikipedia documents~\cite{chen-etal-2023-pre-trained}, with textual queries limited to 128 tokens. As shown in Fig.~\ref{fig:sample}, our constructed dataset is featured by pairs of multimodal queries and passages including responses to different queries about the same image, advancing the capability to retrieve relevant information from multimodal queries. This process yielded a total of 1.36 million QA pairs; further data statistics are provided in Tab.~\ref{tab:data_statistics} and Tab.~\ref{tab:each}.

Table~\ref{tab:data_p} summarizes the zero-shot retrieval performance for each source dataset. The results demonstrate that our conversion process effectively leverages complementary strengths from various datasets, underscoring the robustness of our approach. Additionally, when unifying WiT data, we assigned textual queries by randomly sampling from the following prompts: ``What is the main object?'', ``Identify the subject of this image.'', ``Who or what is the subject in this picture?'', ``Identify the main entity.'', and ``What is the core object or subject shown here?''.

\subsection{Architectural Differences: MIRe vs. PreFLMR}

Our architecture differs from PreFLMR in several important aspects. In PreFLMR, visual tokens serve as queries, and the hidden states of the text encoder act as keys and values within a cross-attention mechanism. Attention scores $\mathcal{A} \in \mathbb{R}^{H \times Q \times K}$ are computed using these components, and value vectors $\mathcal{V}$ are derived from the truncated text encoder hidden states, which may cause information loss. The resulting outputs have dimensions $[H, Q, D]$ and are reshaped to $[Q, H\cdot D]$ before undergoing further reduction by the ColBERT head.

In contrast, MIRe uses textual embeddings $E_t$ from the ColBERT head as queries, with visual tokens split into multiple heads for keys and values. Here, $Q$ corresponds to the number of textual tokens and $K$ to the number of visual tokens, reversing the modality roles compared to PreFLMR. Value vectors are obtained directly from the visual embeddings without truncation. Output aggregation is performed by mean-pooling across the sequence dimension, producing representations of shape $[H, D]$, which are passed through a linear layer without further dimensionality reduction.

% \begin{table*}[!t]
% \centering
% \resizebox{1.\columnwidth}{!}{%
% \begin{tabular}{l}
% \toprule
% What is the main object? \\
% Identify the subject of this image. \\
% Who or what is the subject in this picture? \\
% Identify the main entity. \\
% What is the core object or subject shown here? \\ \bottomrule
% \end{tabular}
% }
% \caption{\textbf{Dummy Prompts.}}
% \end{table*}

\end{document}